\documentclass[english,twocolumn]{svjour3}          
\smartqed  
\usepackage[T1]{fontenc}
\usepackage[utf8]{inputenc}
\usepackage{color}
\usepackage{wrapfig}
\usepackage{mathtools}
\usepackage{url}
\usepackage{amsmath}
\usepackage{amssymb}
\usepackage{graphicx}

\usepackage[breaklinks]{hyperref}
\hypersetup{
	bookmarks=true,         
	colorlinks=true,       
    linkcolor=blue,          
    citecolor=green,        
    filecolor=magenta,      
    urlcolor=magenta           
    }

\makeatletter

\spnewtheorem{thmresult}{Result}{\bf}{\it} 

\DeclareMathOperator{\tr}{trace}

\makeatother

\usepackage{babel}

\global\long\def\R{\mathbb{R}}

\global\long\def\X{\mathbf{X}}

\global\long\def\mA{\mathtt{A}}

\global\long\def\mG{\mathtt{G}}

\global\long\def\mR{\mathtt{R}}

\global\long\def\Rot{\mathtt{R}}

\global\long\def\ba{\mathbf{a}}
\global\long\def\bb{\mathbf{b}}

\global\long\def\bc{\mathbf{c}}

\global\long\def\be{\mathbf{e}}

\global\long\def\bu{\mathbf{u}}

\global\long\def\bv{\mathbf{v}}

\global\long\def\prtl#1#2{\frac{\partial#1}{\partial#2}}

\global\long\def\Cross#1{\left[#1\right]_{\times}}

\newcommand{\bvu}{\mathbf{\bar{\bv}}}
\newcommand{\bvuc}{\bar{v}}
\newcommand{\SpaceSkewMats}{\text{Skew}_{3}}
\newcommand{\soAlgebra}{\mathfrak{so}(3)}
\newcommand{\SOgroup}{\mathcal{SO}(3)}

\journalname{J Math Imaging Vis}

\begin{document}

\title{A compact formula for the derivative of a 3-D rotation in exponential
coordinates%
\thanks{G. Gallego is supported by the Marie Curie - COFUND Programme of the
EU, as part of the Seventh Framework Programme (FP7). A. Yezzi is
supported by National Science Foundation (NSF) Grants: CCF-1347191
and CMMI-1068624. This work has been partially supported by the Spanish
Government under project TEC2010-20412 (Enhanced 3DTV), and by FP7's
project 610691 (BRIDGET).%
}}


\author{Guillermo Gallego \and Anthony Yezzi}


\institute{G. Gallego \at  was with Grupo de Tratamiento de Imágenes, Universidad
Politécnica de Madrid, E-28040, Spain. He is now with Robotics and
Perception Group, AI Laboratory, University of Zürich, 8001 Zürich,
Switzerland.\\
\email{guillermo.gallego@ifi.uzh.ch}\\
\and A. Yezzi \at School of Electrical and Computer Engineering,
Georgia Institute of Technology, Atlanta, Georgia 30332, USA.}

\date{Received: date / Accepted: date}

\maketitle

\begin{abstract}
We present a compact formula for the derivative of a 3-D rotation
matrix with respect to its exponential coordinates. A geometric interpretation
of the resulting expression is provided, as well as its agreement
with other less-compact but better-known formulas. To the best of
our knowledge, this simpler formula does not appear anywhere in the
literature. We hope by providing this more compact expression to alleviate
the common pressure to reluctantly resort to alternative representations
in various computational applications simply as a means to avoid the
complexity of differential analysis in exponential coordinates. 
\end{abstract}

\section{Introduction\label{sec:Introduction}}

Three-dimensional rotations have numerous applications in many scientific
areas, from quantum mechanics to stellar and planetary rotation, including
the kinematics of rigid bodies. In particular, they are widespread
in computer vision and robotics to describe the orientation of cameras
and objects in the scene, as well as to describe the kinematics of
wrists and other parts of a robot or a mobile computing device.

Space rotations have three degrees of freedom, and admit several ways
to represent and operate with them. Each representation has advantages
and disadvantages. Among the most common representations of rotations
are Euler angles, axis-angle representation, exponential coordinates,
unit quaternions, and rotation matrices. Euler angles~\cite[p. 31]{murray1994},
axis-angle and exponential coordinates~\cite[p. 30]{murray1994}
are very easy to visualize because they are directly related to world
models; they are also compact representations, consisting of 3-4 real
numbers. These representations are used as parametrizations of $3\times3$
rotation matrices~\cite[p. 23]{murray1994}, which are easier to
work with but require nine real numbers. Unit quaternions (also known
as Euler-Rodrigues parameters) \cite[p. 33]{altmann1986,murray1994}
are a less intuitive representation, but nevertheless more compact
(4 real numbers) than $3\times3$ matrices, and are also easy to work
with. Historical notes as well as additional references on the representations
of rotations can be found in~\cite[p. 43]{ma2003invitation}, \cite{Bauchau2003}.

In many applications, it is not only necessary to know how to represent
rotations and carry out simple group operations but also to be able
to perform some differential analysis. This often requires the calculation
of derivatives of the rotation matrix, for example, to find optimal
rotations that control some process or that minimize some cost function
(in cases where a closed form solution does not exist)~\cite{Gurwitz1989,Horn1990}.
Such is the case for the optimal pose estimation problem long studied
within the computer vision and photogrammetry communities~\cite{Olsson2006ICPR,HartleyICCV2007,Enqvist2008},
as well as for other related problems~\cite{UnalTPAMI2007,Dambreville2010,KuehnelTR2003,SandhuTPAMI2010}.

Here we consider rotations parametrized by exponential coordinates
using the well-known Euler-Rodrigues formula, and compute a compact
expression, in matrix form, for the derivative of the parametrized
rotation matrix. We also give a geometric interpretation of the formula
in terms of the spatial decomposition given by the rotation axis.
To the authors' knowledge, the result presented here has not been
shown before and it fills a gap in the literature, at an intermediate
point between numerical differentiation, the derivative at the identity
for incremental rotations and general formulas for Lie groups. By
providing this simpler, compact formula for the derivative of the
rotation matrix we hope to alleviate the common pressure to reluctantly
resort to alternative representations or the framework of incremental
rotations (i.e., local charts) in various computational applications
simply as a means to avoid the complexity of differential analysis
in exponential coordinates.

The paper is organized as follows: Section~\ref{sec:Rotation-parametrization}
reviews the theory of 3-D rotations parametrized by exponential coordinates.
Section~\ref{sec:Derivative-of-Rot} presents the main contributions
of this paper, where proofs and secondary results have been moved
to appendices for readers interested in technical details. Finally,
conclusions are given in Section~\ref{sec:Conclusion}.

\section{Parametrization of a rotation\label{sec:Rotation-parametrization}}

In this section, we review the parametrization of space rotations
using exponential coordinates, before proceeding to calculating derivatives. 

A three-dimensional rotation is a circular movement of an object around
an imaginary line called the rotation axis. The rotation angle measures
the amount of circular displacement. Rotations preserve Euclidean
distance and orientation. Algebraically, the rotation of a point $\X=(X,Y,Z)^{\top}$
to a point $\X'=(X',Y',Z')^{\top}$ can be expressed as $\X'=\Rot\X$,
where the rotation matrix $\Rot$ is a $3\times3$ orthogonal matrix
($\Rot^{\top}\Rot=\Rot\Rot^{\top}=\mbox{Id}$, the identity matrix)
with determinant $\det(\Rot)=1$. 

The space of 3-D rotations is known as the matrix Lie group $\SOgroup$
(special orthogonal group of order three)~\cite[p. 24]{murray1994},
and it is not isomorphic to $\R^{3}$~\cite{Stuelpnagel1964}. It
has the structure of both a non-commutative group (under the composition
of rotations) and a manifold for which the group operations are smooth.
Since $\SOgroup$ is a differentiable manifold, each of its points
(i.e., rotations) has a tangent space, and the corresponding vector
space~$\SpaceSkewMats$ consists of all (real) $3\times3$ skew-symmetric
matrices, which can be thought of as infinitesimal rotations~\cite[p. 25]{ma2003invitation}.
Moreover, the exponential map $\exp:\soAlgebra\to\SOgroup$ can be
defined, which allows one to recapture the local group structure of
$\SOgroup$ from the Lie algebra $\soAlgebra$, the latter consisting
of $\SpaceSkewMats$ together with the binary operation (Lie bracket
or commutator) $[A,B]=AB-BA$, with $A,B\in\SpaceSkewMats$.

The Euler-Rodrigues formula~\cite{History1989}\cite[p. 28]{murray1994}
states that the rotation matrix representing a circular movement of
angle $\theta$ (in radians) around a specified axis $\bvu\in\R^{3}$
is given by
\begin{equation}
\Rot=\mbox{Id}+\sin\theta\,\Cross{\bvu}+(1-\cos\theta)\Cross{\bvu}^{2},\label{eq:RodriguesRot1-1}
\end{equation}
where $\bvu$ is a unit vector, and 
\begin{equation}
\Cross{\ba}\coloneqq\left(\begin{array}{ccc}
0 & -a_{3} & a_{2}\\
a_{3} & 0 & -a_{1}\\
-a_{2} & a_{1} & 0
\end{array}\right)\in\SpaceSkewMats\label{eq:DefCrossProdMatrix}
\end{equation}
is the cross product (skew-symmetric) matrix such that $\Cross{\ba}\bb=\ba\times\bb$,
for all $\ba=(a_{1},a_{2},a_{3})^{\top},\bb\in\R^{3}$.

An alternative formula for~\eqref{eq:RodriguesRot1-1} is
\begin{equation}
\Rot=\cos\theta\,\mbox{Id}+\sin\theta\Cross{\bvu}+(1-\cos\theta)\bvu\bvu^{\top}\label{eq:RodriguesRotuut}
\end{equation}
because any unit vector $\bvu$ satisfies
\begin{equation}
\Cross{\bvu}^{2}=\bvu\bvu^{\top}-\mbox{Id}.\label{eq:CrossuSquared}
\end{equation}

The exponential coordinates~\cite[p. 30]{murray1994} given by the
rotation vector $\bv\coloneqq\theta\bvu$ are a natural and compact
representation of the rotation in terms of its geometric building
blocks. They are also called the canonical coordinates of the rotation
group. The Euler-Rodrigues rotation formula~\eqref{eq:RodriguesRot1-1}
is a closed form expression of the aforementioned exponential map~\cite[p. 29]{murray1994}
\begin{equation}
\Rot=\exp(\Cross{\bv})\coloneqq\sum_{k=0}^{\infty}\frac{1}{k!}\Cross{\bv}^{k}=\sum_{k=0}^{\infty}\frac{\theta^{k}}{k!}\Cross{\bvu}^{k}.\label{eq:RExponential}
\end{equation}

 Moreover, since the exponential map considered is surjective, every
rotation matrix can be written as~\eqref{eq:RExponential} for some
coordinates $\bv$, specifically, those with $\|\bv\|\leq\pi$, i.e.,
in the closed ball of radius $\pi$ in $\R^{3}$. Hence, exponential
coordinates can be used either locally (to represent incremental rotations
between two nearby configurations) or globally (to represent total
rotations with respect to a reference one)~\cite{RittoCorrea2002}.
More observations of this parametrization can be found in~\cite[p. 624]{Hartley-Zisserman}.

To retrieve the exponential coordinates or the axis-angle representation
of a rotation matrix, we use the log map, $\log:\SOgroup\to\soAlgebra$,
given in~\cite[p. 27]{ma2003invitation} by
\[
\theta=\|\bv\|=\arccos\left(\frac{\tr(\Rot)-1}{2}\right)
\]
 and, if $\theta\neq0$ and $\Rot_{ij}$ are the entries of $\Rot$,
\[
\bvu=\frac{\bv}{\|\bv\|}=\frac{1}{2\sin\theta}\bigl(\Rot_{32}-\Rot_{23},\Rot_{13}-\Rot_{31},\Rot_{21}-\Rot_{12}\bigr)^{\top}.
\]
Without loss of generality, if $\theta$ is the principal value of
the inverse cosine function then $\bv$ lies within the ball of radius
$\pi$, as assumed from now on. In the identity case, $\Rot=\mbox{Id}$,
then $\theta=0$ and $\bvu$ can be chosen arbitrarily.

\section{Derivative of a rotation\label{sec:Derivative-of-Rot}}

Given the exponential coordinate parametrization~\eqref{eq:RExponential},
we consider the calculation of the derivative of the Rotation matrix,
which is a relevant topic on its own as well as due to its broad range
of applications. Although formulas exist to express the derivative
of the exponential map in general Lie groups~\cite[p. 95]{helgason1962differential}\cite[p. 70]{BHall2004},
they are not computationally friendly. Instead, for the rotation group,
researchers commonly resort to one of the following alternatives:
numerical differentiation, using a complicated analytical formula
for the derivative (see~\eqref{eq:Rodrigues_dRdwi}) or reformulating
the problem using incremental rotations so that formulas for the simplified
case (linearization around the identity element) are used. 

Here, we bridge the gap between the aforementioned general formulas
and alternatives for the rotation group by providing a simple, analytical
and computationally friendly formula to calculate the derivative
of a rotation. We also give the geometric interpretation in terms
of the spatial decomposition according to the rotation axis. After
many false starts, we report the path that lead to the formula using
well-known matrix identities. We conjecture that there is a way to
obtain such formula from the general one for Lie groups, but so far
we have not found it.

The incremental rotation approach has the following explanation.
The Lie group framework allows $\SOgroup$ to be locally replaced
by its linearized version, i.e., the Lie algebra $\soAlgebra$, whose
vector space is the tangent space of $\SOgroup$ at the identity element~\cite[p. 26]{ma2003invitation}.
This element plays a key role in differential analysis with the exponential
map: it shows that rotations may be linearly approximated using three
so-called group generators (the standard basis for~$\SpaceSkewMats$)
\begin{equation}
\mG_{i}\coloneqq\prtl{}{v_{i}}\exp(\Cross{\bv})\Bigr|_{\bv=\mathbf{0}}=\Cross{\be_{i}},\label{eq:Generators}
\end{equation}
where $\bv=(v_{1},v_{2},v_{3})^{\top}$ and $\be_{i}$ is the $i$-th
vector of the standard basis in $\R^{3}$. And it also provides a
means to calculate derivatives of rotations as long as they are written
in an incremental way, e.g., $\Rot=\exp(\Cross{\bv})\Rot_{0}$, so
that derivatives are evaluated at $\bv=\mathbf{0}$, as in~\eqref{eq:Generators}.

Next, let us show a formula for the derivative of a rotation at an
arbitrary element, not necessarily the simplified case of the identity
element of the rotation group. First we show a well-known but complicated
one and then our contribution.

Stemming from the Euler-Rodrigues formula~\eqref{eq:RodriguesRot1-1},
the derivative of a rotation $\Rot(\bv)=\exp(\Cross{\bv})$ with respect
to its exponential coordinates $\bv$ is given by 
\begin{eqnarray}
\prtl{\Rot}{v_{i}} & = & \cos\theta\,\bvuc_{i}\Cross{\bvu}+\sin\theta\,\bvuc_{i}\Cross{\bvu}^{2}+\frac{\sin\theta}{\theta}\Cross{\be_{i}-\bvuc_{i}\bvu}\nonumber \\
 &  & +\frac{1-\cos\theta}{\theta}\left(\be_{i}\bvu^{\top}+\bvu\be_{i}^{\top}-2\bvuc_{i}\bvu\bvu^{\top}\right),\label{eq:Rodrigues_dRdwi}
\end{eqnarray}
where $\theta=\|\bv\|$ and $\bvu=(\bvuc_{1},\bvuc_{2},\bvuc_{3})^{\top}=\bv/\|\bv\|$.
Formula~\eqref{eq:Rodrigues_dRdwi} is used, for example, in the
OpenCV library~\cite{opencvlib} (having more than 50 thousand people
of user community and estimated number of downloads exceeding 7 million)
if the rotation vector $\bv$ is passed as argument to the appropriate
function (\texttt{cvRodrigues}). The proof of~\eqref{eq:Rodrigues_dRdwi}
is given in Appendix~\ref{sec:Derivative-formula-sines}.

Here, however we follow a different approach and first compute the
derivative of the product $\Rot\bu$ where $\bu$ is independent of
the exponential coordinates $\bv$. Once obtained a compact formula,
it is used to compute the derivatives of the rotation matrix itself.

\begin{thmresult}\label{thm:derivative-of-Ru}The derivative of $\Rot(\bv)\bu=\exp(\Cross{\bv})\bu$
with respect to the exponential coordinates $\bv$, where $\bu$ is
independent of $\bv$, is
\begin{equation}
\prtl{\Rot(\bv)\bu}{\bv}=-\Rot\Cross{\bu}\frac{\bv\bv^{\top}+(\Rot^{\top}-\mbox{Id})\Cross{\bv}}{\|\bv\|^{2}}.\label{eq:compact_dRvecdwi}
\end{equation}

\end{thmresult}

The proof is given in Appendix~\ref{sec:Proof-Result-One}.

\subsection{Geometric interpretation}

Let the decomposition of a vector $\bb$ onto the subspaces parallel
and perpendicular components to the rotation axis $\bvu$ be $\bb=\bb_{\parallel}+\bb_{\perp}$
, where $\bb_{\parallel}\propto\bvu$ is parallel to the rotation
axis and $\bb_{\perp}\perp\bvu$ lies in the plane orthogonal to the
rotation axis. Then, observe that formula~\eqref{eq:compact_dRvecdwi}
provides insight about the action of $\partial(\Rot\bu)/\partial\bv$
on a vector $\bb$. Such operation has two components according to
the aforementioned decomposition along/orthogonal to the rotation
axis,
\[
\prtl{\Rot\bu}{\bv}\,\bb=-\Rot\Cross{\bu}\Bigl((\bb_{\parallel}\cdot\bvu)\bvu+\frac{(\Rot^{\top}-\mbox{Id})\Cross{\bvu}\bb_{\perp}}{\|\bv\|}\Bigr),
\]
and notice that both components scale differently: the first term
$(\bb_{\parallel}\cdot\bvu)\bvu$ depends solely on $\bb_{\parallel}$,
whereas the second term involves $\Cross{\bvu}\bb_{\perp}/\|\bv\|$,
which depends on both $\bb_{\perp}$ and $\|\bv\|$. This information
is difficult to extract by using a formula like~\eqref{eq:Rodrigues_dRdwi}.

Another way to look at the geometric interpretation of our formula
is through sensitivity analysis. The first order Taylor series approximation
of the rotated point $\bu'=\Rot(\bv)\bu$ around $\bv$ is
\[
\begin{split} & \bu'(\bv+\delta\bv)\approx\bu'(\bv)+\prtl{\Rot\bu}{\bv}\,\delta\bv\\
 & =\bu'(\bv)-\Rot\Cross{\bu}\Bigl((\delta\bv_{\parallel}\cdot\bvu)\bvu+\frac{(\Rot^{\top}-\mbox{Id})\Cross{\bvu}\delta\bv_{\perp}}{\|\bv\|}\Bigr),
\end{split}
\]
 where $\delta\bv=\delta\bv_{\parallel}+\delta\bv_{\perp}$. As the
rotation $\Rot(\bv)$ is perturbed, there are two different types
of changes:
\begin{itemize}
\item If the perturbation $\delta\bv$ is such that only the amount of rotation
changes, but not the direction of rotation (rotation axis), i.e.,
$\delta\bv_{\perp}=\mathbf{0}$, the rotated point becomes $\bu'(\bv+\delta\bv)\approx\bu'(\bv)-\|\delta\bv\|\,\Rot(\bu\times\bvu)$,
where the change is proportional to the rotation of $\bu\times\bvu$.
Equivalently, using property~\eqref{eq:CrossProdOfTransformedVecs}
with $\mG=\mR$, $\Rot(\bu\times\bvu)=(\Rot\bu)\times(\Rot\bvu)=\Rot\bu\times\bvu$,
the change is perpendicular to both $\Rot\bu$ and the rotation axis
$\bvu$, which is easy to visualize geometrically since the change
is represented by the tangent vector to the circumference traced out
by point $\bu$ as it rotates around the fixed axis $\bvu$, $(\bu'(\bv+\delta\bv)-\bu'(\bv))/\|\delta\bv\|$.
\item If the perturbation $\delta\bv$ is such that only the direction of
the rotation changes, but not the amount of rotation, i.e., $\delta\bv_{\parallel}=\mathbf{0}$,
the rotated point becomes $\bu'(\bv+\delta\bv)\approx\bu'(\bv)-\|\bv\|^{-1}\Rot\Cross{\bu}(\Rot^{\top}-\mbox{Id})\Cross{\bvu}\delta\bv$.
The scaling is different from previous case, since now the change
in $\bu'$ depends on both $\delta\bv_{\perp}$ and $\|\bv\|$.
\end{itemize}
For an arbitrary perturbation, the change on the rotated point has
two components: one due to the part of the perturbation that modifies
the amount of rotation, and another one due to the part of the perturbation
that modifies the direction of the rotation.

\subsection{Compact formula for the derivative of the rotation matrix}

Next, we use Result~\ref{thm:derivative-of-Ru} to compute the derivatives
of the rotation matrix itself with respect to the exponential coordinates~\eqref{eq:RExponential},
without re-doing all calculations.

\begin{thmresult}\label{thm:derivative-of-R}The derivative of $\Rot(\bv)=\exp(\Cross{\bv})$
with respect to its exponential coordinates $\bv=(v_{1},v_{2},v_{3})^{\top}$
is
\begin{equation}
\prtl{\Rot}{v_{i}}=\frac{v_{i}\Cross{\bv}+\Cross{\bv\times(\mbox{Id}-\Rot)\be_{i}}}{\|\bv\|^{2}}\,\Rot,\label{eq:compact_dRdwi}
\end{equation}
where $\be_{i}$ is the $i$-th vector of the standard basis in~$\R^{3}$.

\end{thmresult}

The proof is given in Appendix~\ref{sec:Proof-Result-Two}. To conclude,
we also need to show that the compact formula~\eqref{eq:compact_dRdwi}
is consistent with~\eqref{eq:Rodrigues_dRdwi}. This is demonstrated
in Appendix~\ref{sec:Agreement-between-derivative}.

\subsection{Derivative at the identity. }

Our result, evaluated at the identity element, agrees with the well-known
result about the so-called generators $\mG_{i}$ of the group~\eqref{eq:Generators}.
This can be shown by computing the limit as $\bv\to0$ of~\eqref{eq:compact_dRdwi},
and using the facts that $\lim_{\bv\to\mathbf{0}}\Rot=\mbox{Id}$
and $\lim_{\bv\to\mathbf{0}}(\mbox{Id}-\Rot)/\|\bv\|=-\Cross{\bvu}$,
\begin{eqnarray*}
\lim_{\bv\to\mathbf{0}}\prtl{\Rot}{v_{i}} & \stackrel{\eqref{eq:compact_dRdwi}}{=} & \lim_{\bv\to\mathbf{0}}\Bigl(\bigl(\bvuc_{i}\Cross{\bvu}+\frac{\Cross{\bvu\times(\mbox{Id}-\Rot)\be_{i}}}{\|\bv\|}\bigr)\Rot\Bigr)\\
 & = & \bvuc_{i}\Cross{\bvu}-\Cross{\bvu\times(\Cross{\bvu}\be_{i})}\\
 & = & \Cross{\bvuc_{i}\bvu-\Cross{\bvu}^{2}\be_{i}}\\
 & \stackrel{\eqref{eq:CrossuSquared}}{=} & \Cross{\be_{i}}.
\end{eqnarray*}

\section{Conclusion\label{sec:Conclusion}}

We have provided a compact formula for the derivative of a rotation
matrix in exponential coordinates. The formula is not only simpler
than existing ones but it also has an intuitive interpretation according
to the geometric decomposition that it provides in terms of the amount
of rotation and the direction of rotation. This, together with the
Euler-Rodrigues formula and the fact that exponential coordinates
provide a global chart of the rotation group are supporting arguments
in favor of using such parametrization for the search of optimal rotations
in first-order finite-dimensional optimization techniques. In addition,
the formula can also provide a simple fix for numerical implementations
that are based on the derivative of a linearization of the rotation
matrix in exponential coordinates.

\appendix

\section{Some cross product relations\label{sec:cross-product-props}}

Let us use the dot notation for the Euclidean inner product $\ba\cdot\bb=\ba^{\top}\bb$.
Also, let $\mG$ be a $3\times3$ matrix, invertible when required
so that it represents a change of coordinates in~$\R^{3}$. 
\begin{eqnarray}
\Cross{\ba}\ba & = & \mathbf{0}\label{eq:Crossakernel}\\
\Cross{\ba}\bb & = & -\Cross{\bb}\ba\label{eq:CrossFlipArgs}\\
\Cross{\ba}\Cross{\bb} & = & \bb\ba^{\top}-(\ba\cdot\bb)\mbox{Id}\label{eq:CrossMatCrossMat}\\
\ba\times(\bb\times\bc) & \stackrel{\eqref{eq:CrossMatCrossMat}}{=} & (\ba\cdot\bc)\bb-(\ba\cdot\bb)\bc\label{eq:TripleCrossProd}\\
\Cross{\ba\times\bb} & = & \bb\ba^{\top}-\ba\bb^{\top}\label{eq:CrossMatOfCrossProd}\\
\Cross{\ba\times\bb} & = & \Cross{\ba}\Cross{\bb}-\Cross{\bb}\Cross{\ba}\label{eq:CrossMatOfCrossProdMats}\\
\Cross{(\mG\ba)\times(\mG\bb)} & = & \mG\Cross{\ba\times\bb}\mG^{\top}\nonumber \\
(\mG\ba)\times(\mG\bb) & = & \det(\mG)\mG^{-\top}(\ba\times\bb)\label{eq:CrossProdOfTransformedVecs}\\
\Cross{\ba}\mG+\mG^{\top}\Cross{\ba} & = & \tr(\mG)\Cross{\ba}-\Cross{\mG\ba}\label{eq:TraceAndCrossProds}\\
\Cross{\mG\ba} & = & \det(\mG)\mG^{-\top}\Cross{\ba}\mG^{-1}\nonumber 
\end{eqnarray}

\section{Proof of Result~\ref{thm:derivative-of-Ru}\label{sec:Proof-Result-One}}
\begin{proof}
Four terms result from applying the chain rule to~\eqref{eq:RodriguesRot1-1}
acting on vector $\bu$. Let us use $\theta=\|\bv\|$ and $\bvu=\bv/\|\bv\|$,
then
\begin{eqnarray*}
\prtl{\Rot\bu}{\bv} & = & \sin\theta\,\prtl{\Cross{\bvu}\bu}{\bv}+\Cross{\bvu}\bu\,\prtl{\sin\theta}{\bv}\\
 &  & +(1-\cos\theta)\prtl{\Cross{\bvu}^{2}\bu}{\bv}+\Cross{\bvu}^{2}\bu\,\prtl{(1-\cos\theta)}{\bv}.
\end{eqnarray*}
The previous derivatives are computed next, using some of the cross
product properties listed in Appendix~\ref{sec:cross-product-props}:
\[
\prtl{\Cross{\bvu}\bu}{\bv}\stackrel{\eqref{eq:CrossFlipArgs}}{=}\prtl{(-\Cross{\bu}\bvu)}{\bvu}\prtl{\bvu}{\bv}=-\Cross{\bu}\prtl{\bvu}{\bv},
\]
with derivative of the unit rotation axis vector
\begin{equation}
\prtl{\bvu}{\bv}=\prtl{}{\bv}\left(\frac{\bv}{\|\bv\|}\right)=\frac{1}{\theta}(\mbox{Id}-\bvu\bvu^{\top})\stackrel{\eqref{eq:CrossuSquared}}{=}-\frac{1}{\theta}\Cross{\bvu}^{2}.\label{eq:dudwAsCross}
\end{equation}
Also by the chain rule,
\begin{eqnarray*}
\prtl{\sin\theta}{\bv} & = & \prtl{\sin\theta}{\theta}\prtl{\theta}{\bv}=\cos\theta\,\bvu^{\top},\\
\prtl{(1-\cos\theta)}{\bv} & = & -\prtl{\cos\theta}{\theta}\prtl{\theta}{\bv}=\sin\theta\,\bvu^{\top},
\end{eqnarray*}
 and, applying the product rule twice,
\begin{eqnarray*}
\prtl{\Cross{\bvu}^{2}\bu}{\bv} & \stackrel{\eqref{eq:CrossuSquared}}{=} & \prtl{\bvu(\bvu^{\top}\bu)}{\bv}=\prtl{\bvu}{\bv}(\bvu^{\top}\bu)+\bvu\prtl{(\bvu^{\top}\bu)}{\bv}\\
 & = & \bigl((\bvu^{\top}\bu)\mbox{Id}+\bvu\bu^{\top}\bigr)\prtl{\bvu}{\bv}\\
 & \stackrel{\eqref{eq:dudwAsCross}}{=} & -\frac{1}{\theta}\bigl((\bvu^{\top}\bu)\mbox{Id}+\bvu\bu^{\top}\bigr)\Cross{\bvu}^{2},
\end{eqnarray*}
which can be rewritten as a sum of cross product matrix multiplications
since
\[
\begin{split} & \bigl((\bvu^{\top}\bu)\mbox{Id}+\bvu\bu^{\top}\bigr)\Cross{\bvu}^{2}\\
 & \stackrel{\eqref{eq:Crossakernel}}{=}\bigl((\bvu^{\top}\bu)\mbox{Id}-\bu\bvu^{\top}+\bvu\bu^{\top}-\bu\bvu^{\top}\bigr)\Cross{\bvu}^{2}\\
 & \stackrel{\eqref{eq:CrossMatCrossMat}\,\eqref{eq:CrossMatOfCrossProd}}{=}-\Cross{\bvu}\Cross{\bu}\Cross{\bvu}^{2}+\Cross{\bu\times\bvu}\Cross{\bvu}^{2}\\
 & \stackrel{\eqref{eq:CrossMatOfCrossProdMats}\,\eqref{eq:CrossuSquared}}{=}-2\Cross{\bvu}\Cross{\bu}\Cross{\bvu}^{2}-\Cross{\bu}\Cross{\bvu}.
\end{split}
\]
 Hence, so far the derivative of the rotated vector is 
\[
\begin{split}\prtl{\Rot\bu}{\bv} & =(\cos\theta\Cross{\bvu}+\sin\theta\Cross{\bvu}^{2})\bu\bvu^{\top}+\frac{\sin\theta}{\theta}\Cross{\bu}\Cross{\bvu}^{2}\\
 & \quad+\frac{1-\cos\theta}{\theta}(2\Cross{\bvu}\Cross{\bu}\Cross{\bvu}^{2}+\Cross{\bu}\Cross{\bvu}).
\end{split}
\]

Next, multiply on the left by $\Rot^{\top}$ and use
\begin{equation}
\Rot^{\top}\Cross{\bvu}\stackrel{\eqref{eq:RodriguesRotuut}\,\eqref{eq:Crossakernel}}{=}\cos\theta\,\Cross{\bvu}-\sin\theta\,\Cross{\bvu}^{2}\label{eq:RtransposeCrossu}
\end{equation}
to simplify the first term of $\Rot^{\top}\partial(\Rot\bu)/\partial\bv$,
\begin{equation}
\begin{split} & \Rot^{\top}(\cos\theta\,\Cross{\bvu}+\sin\theta\,\Cross{\bvu}^{2})\bu\bvu^{\top}\\
 & \stackrel{\eqref{eq:RtransposeCrossu}}{=}(\cos^{2}\theta\,\Cross{\bvu}-\sin^{2}\theta\,\Cross{\bvu}^{3})\bu\bvu^{\top}\\
 & \stackrel{\eqref{eq:CrossuSquared}}{=}(\cos^{2}\theta+\sin^{2}\theta)\,\Cross{\bvu}\bu\bvu^{\top}\\
 & \stackrel{\eqref{eq:CrossFlipArgs}}{=}-\Cross{\bu}\bvu\bvu^{\top}.
\end{split}
\label{eq:FirstTermRtransDeriv}
\end{equation}
 For the remaining term of $\Rot^{\top}\partial(\Rot\bu)/\partial\bv$,
we use the transpose of~\eqref{eq:RodriguesRot1-1} and apply $\Cross{\bvu}\Cross{\bu}\Cross{\bvu}\stackrel{\eqref{eq:CrossMatCrossMat}}{=}-(\bu^{\top}\bvu)\Cross{\bvu}$
to simplify
\[
\begin{split} & \bigl(\mbox{Id}-\sin\theta\,\Cross{\bvu}+(1-\cos\theta)\Cross{\bvu}^{2}\bigr)\cdot\bigl(\sin\theta\Cross{\bu}\Cross{\bvu}^{2}\\
 & \quad+(1-\cos\theta)(2\Cross{\bvu}\Cross{\bu}\Cross{\bvu}^{2}+\Cross{\bu}\Cross{\bvu})\bigr)\\
 & =\sin\theta\,\Cross{\bu}\Cross{\bvu}^{2}+(1-\cos\theta)\Cross{\bu}\Cross{\bvu}\\
 & +(\bu^{\top}\bvu)\bigl(-2(1-\cos\theta)+\sin^{2}\theta\,+(1-\cos\theta)^{2}\bigr)\Cross{\bvu}\\
 & =\Cross{\bu}\bigl(\sin\theta\,\Cross{\bvu}^{2}-(1-\cos\theta)\Cross{\bvu}^{3}\bigr)\\
 & =-\Cross{\bu}(\Rot^{\top}-\mbox{Id})\Cross{\bvu},
\end{split}
\]
where the term in $(\bu^{\top}\bvu)$ vanished since $\sin^{2}\theta-2(1-\cos\theta)+(1-\cos\theta)^{2}=0$.
Collecting terms,
\begin{equation}
\Rot^{\top}\prtl{\Rot\bu}{\bv}=-\Cross{\bu}\bigl(\bvu\bvu^{\top}+\frac{1}{\theta}(\Rot^{\top}-\mbox{Id})\Cross{\bvu}\bigr).\label{eq:RtransposeDerivRv}
\end{equation}
Finally, multiply~\eqref{eq:RtransposeDerivRv} on the left by $\Rot$
and use $\Rot\Rot^{\top}=\mbox{Id}$, $\theta=\|\bv\|$, $\bvu=\bv/\|\bv\|$
to obtain~\eqref{eq:compact_dRvecdwi}.
\end{proof}

\section{Proof of Result~\ref{thm:derivative-of-R}\label{sec:Proof-Result-Two}}
\begin{proof}
Stemming from~\eqref{eq:compact_dRvecdwi}, we show that it is possible
to write
\begin{equation}
\prtl{\Rot}{v_{i}}\bu=\mA\bu\label{eq:dRvdwi_operatorform}
\end{equation}
for some matrix $\mA$ and for all vector $\bu$ independent of $\bv$.
Thus in this operator form, $\mA$ is indeed the representation of
$\partial\Rot/\partial v_{i}$. First, observe that
\[
\prtl{\Rot}{v_{i}}\bu=\prtl{}{v_{i}}(\Rot\bu)=\prtl{}{\bv}(\Rot\bu)\,\be_{i},
\]
then substitute~\eqref{eq:compact_dRvecdwi} and simplify using the
cross-product properties to arrive at~\eqref{eq:dRvdwi_operatorform}:
\begin{eqnarray}
\prtl{\Rot}{v_{i}}\bu & = & -\|\bv\|^{-2}\,\Rot\Cross{\bu}\bigl(\bv\bv^{\top}+(\Rot^{\top}-\mbox{Id})\Cross{\bv}\bigr)\be_{i}\nonumber \\
 & = & -\|\bv\|^{-2}\,\Rot\Cross{\bu}\bigl(\bv\bv^{\top}+\Cross{\bv}(\Rot^{\top}-\mbox{Id})\bigr)\be_{i}\nonumber \\
 & = & -\|\bv\|^{-2}\,\Rot\Cross{\bu}\bigl(\bv v_{i}+\bigl(\bv\times(\Rot^{\top}-\mbox{Id})\be_{i}\bigr)\bigr)\nonumber \\
 & \stackrel{\eqref{eq:CrossFlipArgs}}{=} & \|\bv\|^{-2}\,\Rot\Cross{v_{i}\bv+\bigl(\bv\times(\Rot^{\top}-\mbox{Id})\be_{i}\bigr)}\bu.\label{eq:DRudviU_aux}
\end{eqnarray}
After some manipulations, \[\Rot\Cross{v_{i}\bv+\bigl(\bv\times(\Rot^{\top}-\mbox{Id})\be_{i}\bigr)} = \Cross{v_{i}\bv+\bigl(\bv\times(\mbox{Id}-\Rot)\be_{i}\bigr)}\Rot,\]
and so, substituting in~\eqref{eq:DRudviU_aux} and using the linearity
of the cross-product matrix~\eqref{eq:DefCrossProdMatrix}, the desired
formula~\eqref{eq:compact_dRdwi} is obtained.
\end{proof}

\section{Agreement between derivative formulas\label{sec:Agreement-between-derivative}}

Here we show the agreement between~\eqref{eq:Rodrigues_dRdwi} and~\eqref{eq:compact_dRdwi}.
First, use $\theta=\|\bv\|$ and the definition of the unit vector
$\bvu=\bv/\theta$, to write~\eqref{eq:compact_dRdwi} as
\begin{equation}
\prtl{\Rot}{v_{i}}=\bvuc_{i}\Cross{\bvu}\Rot+\frac{1}{\theta}\Cross{\bvu\times(\mbox{Id}-\Rot)\be_{i}}\Rot.\label{eq:dRdwi-u1}
\end{equation}
Using~\eqref{eq:RodriguesRotuut} and $\Cross{\bvu}\bvu=\mathbf{0}$,
it follows that
\[
\Cross{\bvu}\Rot=\cos\theta\,\Cross{\bvu}+\sin\theta\,\Cross{\bvu}^{2}.
\]
 Also, since $\Cross{\bvu\times\bb}=\bb\bvu^{\top}-\bvu\bb^{\top}$
and $\Rot^{\top}\bvu=\bvu$, it yields
\begin{eqnarray*}
\Cross{\bvu\times(\mbox{Id}-\Rot)\be_{i}}\Rot & = & (\mbox{Id}-\Rot)\be_{i}\bvu^{\top}\Rot-\bvu\be_{i}^{\top}(\mbox{Id}-\Rot^{\top})\Rot\\
 & = & (\mbox{Id}-\Rot)\be_{i}\bvu^{\top}-\bvu\be_{i}^{\top}(\Rot-\mbox{Id})\\
 & = & \be_{i}\bvu^{\top}+\bvu\be_{i}^{\top}-(\Rot\be_{i}\bvu^{\top}+\bvu\be_{i}^{\top}\Rot),
\end{eqnarray*}
and expanding $\Rot\be_{i}$ and $\be_{i}^{\top}\Rot$ in the previous
formula by means of~\eqref{eq:RodriguesRotuut}, we obtain
\[
\begin{split} & \Cross{\bvu\times(\mbox{Id}-\Rot)\be_{i}}\Rot\\
 & =\be_{i}\bvu^{\top}+\bvu\be_{i}^{\top}\\
 & \quad-\bigl(\cos\theta\,\be_{i}\bvu^{\top}+\sin\theta\,\Cross{\bvu}\be_{i}\bvu^{\top}+(1-\cos\theta)\bvuc_{i}\bvu\bvu^{\top}\bigr)\\
 & \quad-\bigl(\cos\theta\,\bvu\be_{i}^{\top}+\sin\theta\,\bvu\be_{i}^{\top}\Cross{\bvu}+(1-\cos\theta)\bvuc_{i}\bvu\bvu^{\top}\bigr)\\
 & =(1-\cos\theta)(\be_{i}\bvu^{\top}+\bvu\be_{i}^{\top}-2\bvuc_{i}\bvu\bvu^{\top})\\
 & \quad-\sin\theta\,(\Cross{\bvu}\be_{i}\bvu^{\top}+\bvu\be_{i}^{\top}\Cross{\bvu}).
\end{split}
\]
Using property~\eqref{eq:TraceAndCrossProds} with $\ba=\bvu$, $\mG=\be_{i}\bvu^{\top}$
we have that
\begin{eqnarray*}
\Cross{\bvu}\be_{i}\bvu^{\top}+\bvu\be_{i}^{\top}\Cross{\bvu} & = & \tr(\be_{i}\bvu^{\top})\Cross{\bvu}-\Cross{\be_{i}\bvu^{\top}\bvu},\\
 & = & \tr(\bvu^{\top}\be_{i})\Cross{\bvu}-\Cross{\be_{i}\|\bvu\|^{2}}\\
 & = & \bvuc_{i}\Cross{\bvu}-\Cross{\be_{i}}\\
 & = & \Cross{\bvuc_{i}\bvu-\be_{i}}.
\end{eqnarray*}
Finally, substituting previous results in~\eqref{eq:dRdwi-u1}, the
desired result~\eqref{eq:Rodrigues_dRdwi} is obtained.

\section{Derivative formula with sines and cosines\label{sec:Derivative-formula-sines}}

Here, we show how to obtain formula~\eqref{eq:Rodrigues_dRdwi}.
First, differentiate the Euler-Rodrigues rotation formula~\eqref{eq:RodriguesRotuut}
with respect to the $i$-th component of the parametrizing vector
$\bv=\theta\bvu$ and take into account that
\[
\theta^{2}=\|\bv\|^{2}\implies\prtl{\theta}{v_{i}}=\frac{v_{i}}{\theta}\eqqcolon\bvuc_{i}.
\]
Applying the chain rule to~\eqref{eq:RodriguesRotuut}, gives 
\begin{eqnarray}
\prtl{\Rot}{v_{i}} & = & -\sin\theta\,\bvuc_{i}\mbox{Id}+\cos\theta\,\bvuc_{i}\Cross{\bvu}+\sin\theta\,\bvuc_{i}\bvu\bvu^{\top}\nonumber \\
 &  & +\sin\theta\,\prtl{\Cross{\bvu}}{v_{i}}+(1-\cos\theta)\prtl{(\bvu\bvu^{\top})}{v_{i}}.\label{eq:RodriguesDeriv1}
\end{eqnarray}
Next, we use
\begin{equation}
\prtl{}{v_{i}}\left(\frac{v_{j}}{\|\bv\|}\right)=\begin{cases}
-\frac{1}{\theta}\bvuc_{i}\bvuc_{j} & i\neq j\\
\frac{1}{\theta}(1-\bvuc_{i}^{2}) & i=j
\end{cases},\label{eq:Dvjdtheta_aux}
\end{equation}
to simplify one of the terms in~\eqref{eq:RodriguesDeriv1}, 
\[
\prtl{\Cross{\bvu}}{v_{i}}=\prtl{}{v_{i}}\Cross{\frac{\bv}{\|\bv\|}}\stackrel{\eqref{eq:Dvjdtheta_aux}}{=}\frac{1}{\theta}\Cross{\be_{i}-\bvuc_{i}\bvu}.
\]
Applying the product rule to the last term in~\eqref{eq:RodriguesDeriv1}
and using
\begin{equation}
\prtl{\bvu}{v_{i}}=\prtl{}{v_{i}}\left(\frac{\bv}{\|\bv\|}\right)=\frac{1}{\theta}(\mbox{Id}-\bvu\bvu^{\top})\be_{i},\label{eq:dudwi_aux}
\end{equation}
gives
\[
\prtl{(\bvu\bvu^{\top})}{v_{i}}\stackrel{\eqref{eq:dudwi_aux}}{=}\frac{1}{\theta}\left(\be_{i}\bvu^{\top}+\bvu\be_{i}^{\top}-2\bvuc_{i}\bvu\bvu^{\top}\right).
\]
Finally, substituting the previous expressions in~\eqref{eq:RodriguesDeriv1},
yields~\eqref{eq:Rodrigues_dRdwi}. A similar proof is outlined in~\cite{KuehnelTR2003}
using Einstein summation index notation.

\begin{acknowledgements}
The authors thank the anonymous reviewers for their helpful comments
and suggestions.
\end{acknowledgements}


\end{document}